\documentclass[sigconf]{acmart}

\AtBeginDocument{%
  }

\setcopyright{acmlicensed}
\copyrightyear{2018}
\acmYear{2018}
\acmDOI{XXXXXXX.XXXXXXX}

\acmConference[Conference acronym 'XX]{Make sure to enter the correct
  conference title from your rights confirmation emai}{June 03--05,
  2018}{Woodstock, NY}
\acmISBN{978-1-4503-XXXX-X/18/06}

\usepackage{multirow} 
\usepackage{pdflscape}
\usepackage[ruled,vlined]{algorithm2e}
\usepackage{enumitem}
\usepackage{listings}
\usepackage{xcolor}

\begin{document}

\title{Conditional Quantile Estimation for Uncertain Watch Time in Short-Video Recommendation}

\author{Chengzhi Lin}
\affiliation{%
  \institution{Kuaishou Technology}
  \city{Beijing}
  \country{China}
}
\email{1132559107@qq.com}

\author{Shuchang Liu}
\affiliation{%
  \institution{Kuaishou Technology}
  \city{Beijing}
  \country{China}
}
\email{liushuchang@kuaishou.com}

\author{Chuyuan Wang}
\affiliation{%
  \institution{Kuaishou Technology}
  \city{Beijing}
  \country{China}
}
\email{wangchuyuan@kuaishou.com}

\author{Yongqi Liu}
\affiliation{%
  \institution{Kuaishou Technology}
  \city{Beijing}
  \country{China}
}
\email{liuyongqi@kuaishou.com}

\renewcommand{\shortauthors}{Trovato et al.}

\begin{abstract}
Accurately predicting watch time is crucial for optimizing recommendations and user experience in short video platforms. However, existing methods that estimate a single average watch time often fail to capture the inherent uncertainty in user engagement patterns. In this paper, we propose Conditional Quantile Estimation (CQE)  to model the entire conditional distribution of watch time. Using quantile regression, CQE characterizes the complex watch-time distribution for each user-video pair, providing a flexible and comprehensive approach to understanding user behavior. We further design multiple strategies to combine the quantile estimates, adapting to different recommendation scenarios and user preferences. Extensive offline experiments and online A/B tests demonstrate the superiority of CQE in watch-time prediction and user engagement modeling.  Specifically, deploying CQE online on a large-scale platform with hundreds of millions of daily active users has led to substantial gains in key evaluation metrics, including active days,  engagement time, and video views.
These results highlight
the practical impact of our proposed approach in enhancing the
user experience and overall performance of the short video recommendation system.
The primary code is available at https://github.com/justopit/CQE.
\end{abstract}




\received{20 February 2007}
\received[revised]{12 March 2009}
\received[accepted]{5 June 2009}

\maketitle

\section{Introduction}
The rapid growth of online video platforms has revolutionized the way users consume digital content, and short videos are emerging as one of the most popular formats. Recommender systems play a crucial role in these platforms by providing personalized content recommendations to enhance user engagement and satisfaction \cite{video1, video2, video3, video4,micro_MARNET,micro_len,DCN,DCNV2,yang2024swat}. Unlike traditional recommendation problems (e.g., e-commerce and news recommendation), a key metric for measuring user interest and engagement in short video recommendation is watch time, which comprehensively reflects users' preferences and involvement. Therefore, accurately predicting watch time is vital for optimizing recommendation strategies and improving user experience.

However, predicting watch time remains a challenging task due to the inherent uncertainty and heterogeneity in user behavior. In real-world scenarios, it is often infeasible to obtain multiple observations of watch time for the same user-video pair under identical conditions, as users rarely watch the same video multiple times in the exact same context. This limitation prevents us from directly estimating the true conditional watch time distribution from the data.

Existing methods \cite{D2Q, CVRDD, WTG, D2Co, TPM, DML, CRAED} often focus on predicting the conditional expectation of watch time, overlooking the complexity and diversity of the conditional watch time distribution. These methods fail to fully capture the behavioral differences across various user-video pairs, leading to limitations in recommendation performance. The inadequacy of using a single average value to characterize these complex distribution patterns underscores the need to model the entire conditional distribution of watch time.

\begin{figure*}[t]
    \centering
    \includegraphics[width=1.0\linewidth]{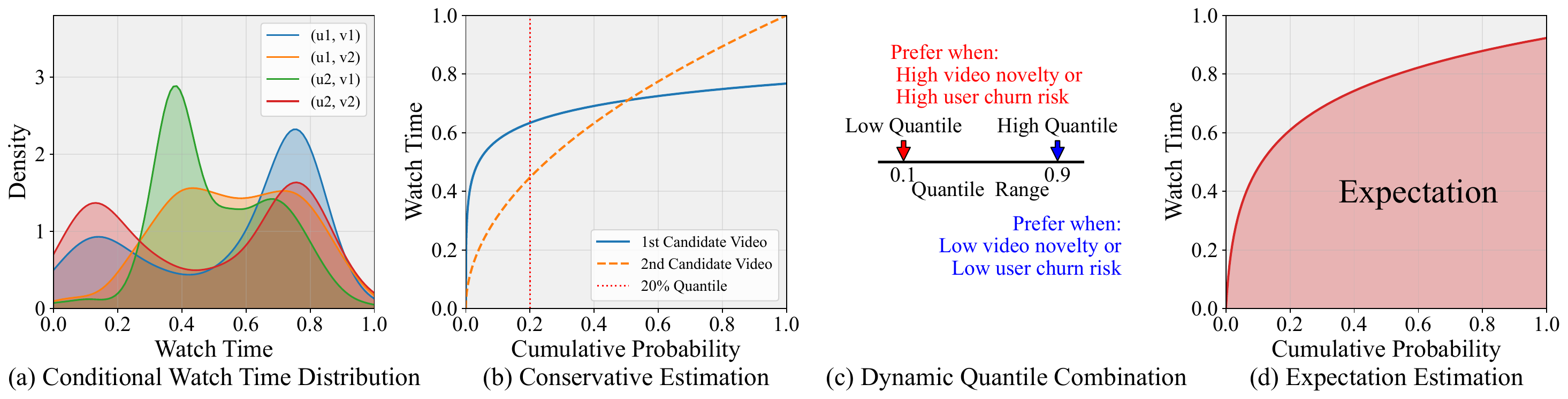}
    \vspace{-0.6cm}
    \caption{Illustration of the conditional watch time distribution and some personalized recommendation strategies. (a) Conditional watch time distributions predicted by our CQE model for different user-video pairs, showcasing the heterogeneity and complexity of user engagement patterns. (b) Conservative estimation strategy  prioritizing user satisfaction by selecting videos with higher lower quantiles when the expected watch times are similar. (c) Dynamic quantile combination strategy adapting to user churn risk or video novelty, employing lower quantiles for high-churn-risk users or unfamiliar videos, and higher quantiles for low-churn-risk users or familiar videos. (d) Expectation estimation strategy providing a global optimization perspective by considering the entire watch time distribution. We normalize watch time by limiting it to a maximum of 300 seconds and scaling it as min(watch time, 300) / 300.}
    \label{fig: motivation}
\end{figure*}

\begin{figure*}[t]
    \centering
    \includegraphics[width=1.0\linewidth]{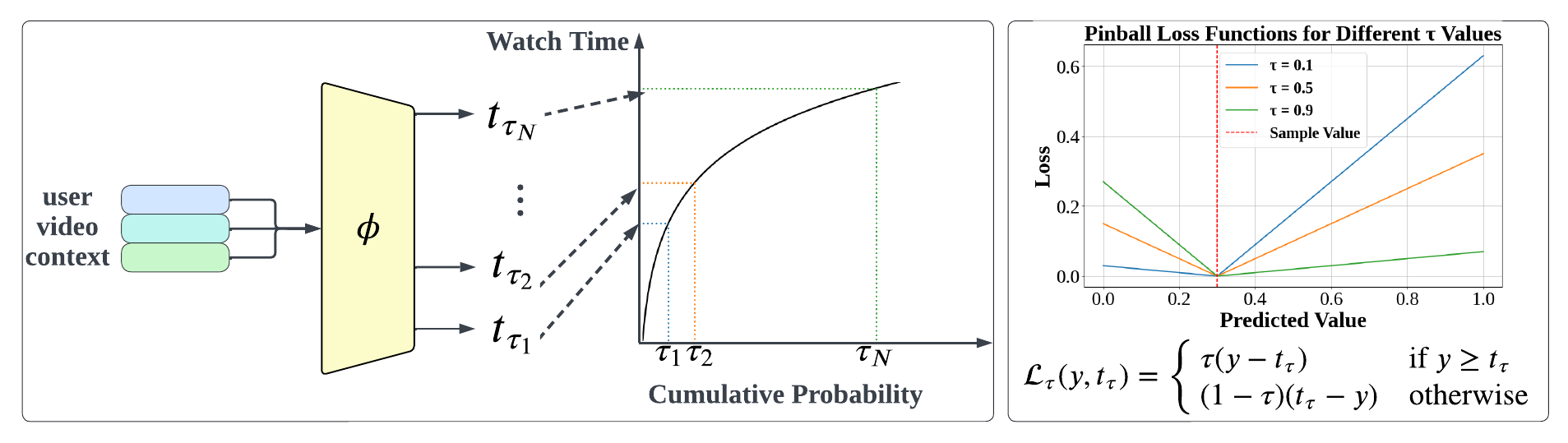}
    \caption{Illustration of the proposed Conditional Quantile Estimation (CQE) model and its training loss. Left: The CQE model architecture, which takes user, video, and context features as input and outputs multiple quantile estimates of watch time. Right: The pinball loss function used for training the model, showing its asymmetric nature for different quantile levels ($\tau$), allowing the model to learn robust quantile estimates across the entire distribution.}
    \label{fig: cqe_framework}
\end{figure*}

To address these challenges, we propose the Conditional Quantile Estimation (CQE) model, which predicts the conditional distribution of watch time given a user-video pair and its associated context. As illustrated in Figure \ref{fig: cqe_framework}, CQE leverages quantile regression—a widely adopted method in statistics~\cite{qr_sta}, econometrics~\cite{qr_eco}, reinforcement learning~\cite{qr_rl}, and large language models~\cite{qr_rm}. By estimating multiple quantiles of the conditional watch time distribution, CQE provides a comprehensive view of potential user engagement patterns.  As illustrated in Figure \ref{fig: motivation}(a), the conditional watch time distributions predicted by our CQE model for different user-video pairs exhibit significant diversity in their shapes, peak positions, and dispersion levels. This heterogeneity reflects the intrinsic uncertainty and variability in user preferences and engagement across different contexts.

Capturing the conditional distribution of watch time is essential for understanding user engagement patterns and developing effective recommendation strategies. By considering the detailed characteristics of the watch time distribution, we can gain insight into the diverse watching behaviors of different user groups. This granular understanding enables us to tailor recommendation strategies for various scenarios and user preferences.

Based on the CQE model, we design three primary recommendation strategies. The conservative estimation strategy (Figure \ref{fig: motivation}(b)) prioritizes user satisfaction by selecting videos with higher lower quantiles when the expected watch times are similar, mitigating disengagement risks for users. The dynamic quantile combination strategy (Figure \ref{fig: motivation}(c))  adapts the choice of quantiles based on factors such as the user's churn risk or video novelty. It assigns more weight to the Low Quantile for high-churn-risk users or novel videos, ensuring a satisfactory experience, and more weight to the High Quantile for low-churn-risk users or familiar videos, potentially offering more engaging recommendations. Finally, the expectation estimation strategy (Figure \ref{fig: motivation}(d)) provides a global optimization perspective, aiming to maximize overall user engagement considering the entire watch time distribution. The diversity of these strategies allows our recommendation system to adapt to different scenarios and user needs, enhancing the quality of personalized recommendations and user experience.

Beyond the heterogeneity shown in Figure 1(a), our empirical analysis on real-world online data quantifies this distributional variability using interquartile range (IQR), defined as $\frac{1}{N/2} \sum_{i=1}^{N/2} t_{\tau_{N-i+1}} - t_{\tau_i}$. This analysis reveals a clear relationship between prediction accuracy and distributional spread, with narrower distributions (IQR in $[0.0,0.1]$) achieving significantly higher prediction accuracy (UAUC: $0.721$, MAE: $0.105$) compared to wider distributions (IQR in $[0.5,1.0]$) with poorer performance (UAUC: $0.536$, MAE: $0.305$). This data-driven evidence from production environments strongly supports the need for our conditional quantile approach that adapts to varying degrees of prediction uncertainty.

To this end, we summarize our key contributions in this paper as follows:
\begin{itemize}[leftmargin=*]
\item We propose the CQE model, which employs the quantile regression technique to model the conditional distribution of watch time in short video recommendation, offering a principled approach to capture the uncertainty in user behavior.
\item We design multiple strategies to combine the quantile estimates from CQE, adapting to different recommendation scenarios and user preferences, enhancing the personalization and diversity of the recommendation system.
\item We validate CQE's superior performance in watch time prediction and user interest modeling through rigorous offline experiments and large-scale online A/B testing, achieving significant improvements in key metrics including active days, engagement duration, and video views.

\end{itemize}

\section{Related Work}\label{sec: related_work}

\subsection{Video Recommendation and Watch-time Prediction}\label{sec: sec: related_work_reco}

Video recommendation systems have evolved to cater to the growing demand for personalized content delivery. With the advent of online video platforms such as YouTube and TikTok, the importance of accurate video recommendation has been underscored by the significant impact on user retention and satisfaction~\cite{video1, video2, video3, video4, CVRDD,DML,CWM,qin2023learning,bae2023competition}. In the realm of video recommendation systems, accurately predicting user engagement through watch time is a critical challenge. Watch time serves as a key metric for gauging user interest and engagement with recommended videos. The initial study~\cite{youtube} focused on improving video recommendations for the YouTube platform, introducing the weighted logistic regression (WLR) technique to predict watch time. 
% This approach has since been recognized as an advanced method in its field. Nevertheless, the WLR's applicability is not straightforward to full-screen video recommendation systems, and it could encounter significant bias problems attributable to its weighted calculation system. 
D2Q~\cite{D2Q} mitigates the duration bias through the implementation of backdoor adjustments and modeling the watch-time quantile under different duration groups. $\text{D}^2$Co \cite{D2Co}  uses a model that corrects both duration bias and noisy watching, providing a more accurate measure of user interest. DVR \cite{WTG} introduces a new metric called WTG (Watch Time Gain) and uses adversarial learning to learn unbiased user preferences. 

Unlike existing bucket-based approaches that discretize watch time and focus solely on predicting conditional expectations (e.g., TPM~\cite{TPM}, CRAED~\cite{CRAED}), our CQE method employs quantile regression to model the complete continuous distribution. This preserves fine-grained distributional information and captures complex patterns, especially in skewed scenarios where tail behavior is crucial. The continuous nature of CQE enables more flexible inference strategies, including Conservative Estimation and Dynamic Quantile Combination. 

Our methodology seamlessly integrates with most duration debiasing techniques, significantly enhancing their predictive performance (Section \ref{sec:exp_backbone}). Watch-time prediction is fundamentally challenged by duration bias \cite{D2Co, D2Q, WTG, TPM}, where users inherently spend more time on longer videos, skewing average watch time toward extended content. This preference complicates accurate user engagement prediction, making debiasing essential.

\subsection{Quantile Regression}\label{sec: related_work_quantile}

Quantile regression is a type of regression method widely used in statistics~\cite{qr_sta}, econometrics~\cite{qr_eco}, reinforcement learning~\cite{qr_rl}, and large language models~\cite{qr_rm}. 
Unlike traditional linear regression, which focuses on estimating the average outcome, quantile regression seeks to estimate the conditional median and other quantiles of the random variable.
This flexible feature provides a more comprehensive understanding of the variable's distributional effects that linear regression may overlook \cite{QR_A1, QR_A2, QR_A3}.
In the context of machine learning, quantile regression has been extended beyond linear models. 
Representative approaches \cite{QB_DL3, QR_DL1, QR_DL2} integrate quantile regression into neural networks, offering means to forecast conditional quantiles in nonlinear and high-dimensional settings. 
QRF \cite{QRF} further deploys quantile regression within random forests, further exemplifying its adaptability and the enhancement of predictive capabilities across diverse models.

The solution presented in this paper intends to integrate the principles of quantile regression into the domain of video recommendation systems. 
By adapting this  approach to account for the uncertainty and variability in watch time, we propose a novel application that enhances the predictive performance of recommendation systems. 
This advancement promotes a more nuanced understanding of user engagement, driving toward more personalized and satisfactory user experiences.

\section{Method}\label{sec: method}

% \subsection{Problem Formulation}\label{sec: method_problem_formulation}

\subsection{Problem Formulation}

In video recommendation systems, our primary objective is to predict user engagement, typically measured by watch time. Let $(u, v)$ denote a user-video pair under context $c$. We define a feature mapping function $\psi(u, v, c)$ that extracts an $n$-dimensional feature vector $\mathbf{x} \in \mathbb{R}^n$. This vector encapsulates user characteristics, video attributes, contextual information, and historical interaction data.

Let $W$ be a random variable representing the watch time. Our goal is to estimate the probability distribution of $W$ conditioned on the input features:

\begin{equation}
    P(W | \mathbf{x}) = P(W | \psi(u, v, c)).
\end{equation}

Unlike traditional approaches \cite{D2Co,WTG,youtube,D2Q,CVRDD, CRAED, TPM} that focus on estimating the conditional expected watch time $\mathbb{E}[W|\mathbf{x}]$, we aim to characterize the entire conditional distribution. This allows us to capture the inherent uncertainty and variability in user engagement patterns, providing a more comprehensive understanding of potential user behaviors.

% The objective of the recommender system is intrinsically a predictive task, where our aim is to estimate the degree of user engagement with the recommended content.
% In the video recommendation scenario, this engagement is mainly estimated by the user's watch-time---the time a user spends watching the video.
% % Hence, we model the problem as a conditional probability challenge, where 
% Formally, we represent each user-video pair as $(u, v)$ under some context $c$, and assume a pre-defined mapping function that extracts an $n$-dimensional feature vector $\mathbf{x} = \psi(u, v, c) \in \mathbb{R}^n$.
% These features encompass contextual information, video characteristics, user profile attributes, and the user's historical interaction data.
% Considering the user's watch-time as a random variable $W$, the \textbf{goal} is to estimate the probability distribution of $W$.

\subsection{Conditional Quantile Estimation Model}

To capture the full distribution of watch time, we propose a Conditional Quantile Estimation (CQE) model. As shown in the left part of Figure \ref{fig: cqe_framework}, this approach allows us to estimate multiple quantiles of the watch time distribution simultaneously, providing a more comprehensive view of potential user engagement.

Let $\{\tau_1, \tau_2, ..., \tau_N\}$ be a set of $N$ predefined quantile levels, where $\tau_i = \frac {i} {N+1}$. 
 Our CQE model aims to estimate the corresponding watch time values $\{t_{\tau_1}, t_{\tau_2}, ..., t_{\tau_N}\}$ for each quantile level, given the input features $\mathbf{x}$:

\begin{equation}
    \{t_{\tau_1}, t_{\tau_2}, ..., t_{\tau_N}\} = \phi(\mathbf{x}; \theta),
\end{equation}
where $\phi(\cdot)$ is a neural network parameterized by $\theta$. To ensure the monotonicity of quantile estimates, we implement the following architecture:
\begin{equation}
\begin{split}
    \mathbf{h} &= f(\mathbf{x}; \theta_f), \\
    \mathbf{d} &= \text{ReLU}(g(\mathbf{h}; \theta_g)), \\
    t_{\tau_i} &= \sum_{j=1}^i d_j, \quad \text{for } i = 1,\ldots,N.
\end{split}
\end{equation}

Here, $f(\cdot)$ and $g(\cdot)$ are neural network components, $\mathbf{h}$ is an intermediate hidden representation, and $\mathbf{d}$ is a vector of nonnegative elements. The final quantile estimates $t_{\tau_i}$ are obtained through cumulative summation, naturally enforcing the ordering constraint $t_{\tau_1} \leq t_{\tau_2} \leq ... \leq t_{\tau_N}$.

This formulation allows our model to capture complex, non-linear relationships between input features and conditional watch time quantiles, while maintaining the necessary monotonicity property of quantile functions.

To provide a clear overview of our CQE approach, we present the pseudocode algorithm in Algorithm \ref{alg:cqe}.

\begin{algorithm}
\SetAlgoLined
\caption{Conditional Quantile Estimation (CQE)}
\label{alg:cqe}

\SetKwProg{Fn}{Function}{:}{}
\SetKwFunction{Train}{Train}
\SetKwFunction{Infer}{Infer}

\Fn{\Train{$D, N, \alpha$}}{
    \KwIn{Training data $D = \{(\mathbf{x}_i, y_i)\}$, number of quantiles $N$, learning rate $\alpha$}
    \KwOut{Trained CQE model $\phi$}
    Initialize model parameters $\theta$\;
    Define quantile levels $\tau_1, ..., \tau_N$\;
    \For{each epoch}{
        \For{each mini-batch $B \subset D$}{
            Compute quantile estimates $\{t_{\tau_1}, ..., t_{\tau_N}\} = \phi(\mathbf{x}; \theta)$\;
            Compute loss $L_{QR}$ using Eq. \ref{eqn:QR} \;
            Update $\theta \gets \theta - \alpha\nabla L_{QR}$\;
        }
    }
    \Return{$\phi$}
}

\Fn{\Infer{$\phi, \mathbf{x}, S$}}{
    \KwIn{Trained CQE model $\phi$, feature vector $\mathbf{x}$, inference strategy $S$}
    \KwOut{Predicted watch time $\hat{y}$}
    Compute quantile estimates $\{t_{\tau_1}, ..., t_{\tau_N}\} = \phi(\mathbf{x})$\;
    Apply inference strategy $S$ to $\{t_{\tau_1}, ..., t_{\tau_N}\}$ to get $\hat{y}$\;
    \Return{$\hat{y}$}
}
\end{algorithm}

The computational complexity of the CQE model is comparable to traditional point estimation methods \cite{D2Co, WTG}, with only a marginal increase due to the estimation of multiple quantiles. In large-scale recommendation systems, the number of unique users and items often reaches hundreds of millions or even billions. These users and items are typically represented by high-dimensional embeddings retrieved using their respective IDs. The number of quantiles required for effective estimation is typically around 100. Therefore, the additional computational cost of CQE is negligible compared to the massive computations required to process the massive user and video features.

\subsection{Training Objective}

To train our CQE model effectively, we employ the pinball loss function \cite{OR}, which is tailored for quantile regression tasks. For a single quantile level $\tau$, the pinball loss is defined as:

\begin{equation}\label{eq:pinball}
    \mathcal{L}_{\tau}(y, t_{\tau}) = 
    \begin{cases}
        \tau(y - t_{\tau}) & \text{if } y \geq t_{\tau}; \\
        (1-\tau)(t_{\tau} - y) & \text{otherwise};
    \end{cases}
\end{equation}
where $y$ is the actual watch time and $t_{\tau}$ is the predicted $\tau$-th quantile.

As illustrated in the right part of Figure \ref{fig: cqe_framework}, the pinball loss function has several key properties:

\begin{enumerate}
    \item Asymmetry: When $\tau \neq 0.5$, the loss is asymmetric around the sample value $y$, with the degree of asymmetry determined by $\tau$.
    \item Linearity: The loss increases linearly with the distance between the predicted and actual values, but with different slopes on each side of $y$.
    \item Quantile-specific penalties: For $\tau > 0.5$, overestimation is penalized more heavily than underestimation, and vice versa for $\tau < 0.5$.
\end{enumerate}

These properties make the pinball loss particularly suitable for quantile estimation. For our multi-quantile model, we aggregate the pinball losses across all quantile levels:

\begin{equation} \label{eqn:QR}
    \mathcal{L}_{\text{QR}} = \sum_{i=1}^N \mathcal{L}_{\tau_i}(y, t_{\tau_i}).
\end{equation}

This aggregated loss function encourages the model to learn accurate quantile estimates across the entire distribution, capturing the full spectrum of potential watch times for each user-video pair.

% \subsection{Conditional Quantile Estimation}\label{sec: method_cqe}

\subsection{Inference Strategies}

Once we have trained our CQE model to estimate multiple quantiles of the conditional watch time distribution, we can employ various strategies for inference. We propose three main approaches: Conservative Estimation, Dynamic Quantile Combination, and Conditional Expectation. Each strategy offers different advantages and is suitable for specific recommendation scenarios.

\subsubsection{Conservative Estimation}

In environments where user satisfaction is paramount and the cost of overestimation is high, we adopt a Conservative Estimation (CSE) strategy. This approach focuses on the lower quantiles of the watch time distribution to ensure a satisfactory user experience.

As illustrated in Figure \ref{fig: motivation}(b), when the expected watch times are similar, we prioritize user satisfaction by selecting videos with higher lower quantiles. This strategy helps mitigate the risks of user disengagement.

Formally, with existing expectation estimation in our online system, we select a lower quantile $\tau_{\text{low}}$ (e.g. $\tau_{\text{low}} = 0.25$) and use its corresponding watch time prediction:

\begin{equation}
    \hat{y}_{\text{CSE}} = t_{\tau_{\text{low}}}.
\end{equation}

This strategy helps mitigate the risk of user disappointment due to overly optimistic recommendations, as the actual watch time is likely to exceed this conservative estimate.

\subsubsection{Dynamic Quantile Combination}

To adapt to varying user preferences and content characteristics, we propose a Dynamic Quantile Combination (DQC) strategy. This approach combines predictions from different quantiles based on contextual factors.

As shown in Figure \ref{fig: motivation}(c), the DQC strategy adapts the choice of quantiles based on the user's churn risk or video novelty. It assigns more weight to the Low Quantile for high-churn-risk users or novel videos, ensuring a satisfactory experience, and more weight to the High Quantile for low-churn-risk users or familiar videos, potentially offering more engaging recommendations. This dynamic approach allows the system to balance between safe recommendations and potentially more rewarding ones based on the user's current state and content familiarity.

Let $k \in [0, 1]$ be a context-dependent blending parameter, we compute the final prediction as:

\begin{equation}
    \hat{y}_{\text{DQC}} = k \cdot t_{\tau_{\text{low}}} + (1-k) \cdot t_{\tau_{\text{high}}},
\end{equation}
where $t_{\tau_{\text{low}}}$ and $t_{\tau_{\text{high}}}$ represent conservative and optimistic quantile predictions, respectively.

The blending parameter $k$ can be adjusted based on factors such as user risk profile, video novelty, or platform objectives. For instance, we might use a higher $k$ (favoring conservative estimates) for new users or novel content, and a lower $k$ for established users or familiar content types.

\subsubsection{Conditional Expectation}

For scenarios in which our goal is to optimize the expected watch time, we employ a conditional expectation strategy. This approach estimates the mean watch time by interpolating between the predicted quantiles.

As depicted in Figure \ref{fig: motivation}(d), the Conditional Expectation Estimation (CDE) strategy provides a global optimization perspective, aiming to maximize overall user engagement considering the entire watch time distribution.

As shown in the left part of Figure \ref{fig: cqe_framework}, these output watch-time values exemplify the watch-time distribution. To recover the mean estimation through the conditional expectation, we face the challenge of not having output values for $\tau \in (\tau_i, \tau_{i+1})$ between any two consecutive quantiles. To circumvent this lack of information, we use the interpolation method to approximate the conditional distribution.

We adopt a linear interpolation between consecutive quantiles, so the expected watch time between $\tau_i$ and $\tau_{i+1}$ becomes $(t_{\tau_i} + t_{\tau_{i+1}})/2(N+1)$. For the two endpoints, we assume $t_0 = t_{\tau_1}$ and $t_1 = t_{\tau_N}$. Then, we can approximate the overall watch-time expectation as:

\begin{equation}
\begin{split}
\hat y_{\text{CDE}} = & \frac{1}{2(N+1)}[(t_{\tau_1} + t_{\tau_1}) + (t_{\tau_1} + t_{\tau_2}) + (t_{\tau_2} + t_{\tau_3})... \\
& +(t_{\tau_{N-2}} + t_{\tau_{N-1}}) + (t_{\tau_{N-1}} + t_{\tau_N}) + (t_{\tau_N} + t_{\tau_N})] \\
= & \frac{1}{N+1}\sum_{i=1}^N t_{\tau_i} + \frac{t_{\tau_1} + t_{\tau_N}}{2(N+1)}.
\end{split}
\end{equation}

Theoretically, this expectation gives the most accurate prediction in general and it would achieve the optimal prediction when $N \to \infty$. Empirically, we will verify its superiority with our experimental analysis in section \ref{sec:offline}. Yet, we remind readers that this strategy may not be well-suited for scenarios where users are not tolerant of bad recommendations or the recommender system requires dynamic controls.

Each of these inference strategies offers unique benefits, allowing the recommendation system to adapt to different objectives and user contexts. Using the rich information provided by our CQE model, we can make more informed and flexible recommendation decisions.

\section{Experiments and Results}\label{sec: experiments}

In this section, we present a comprehensive evaluation of our Conditional Quantile Estimation (CQE) model through both online A/B tests and offline experiments. Our experimental design aims to address several interconnected research questions: 
\begin{itemize}[leftmargin=*]
\item RQ1: How do different CQE strategies perform in real-world scenarios?

\item RQ2: How does CQE compare with state-of-the-art methods in watch time prediction and user interest modeling?

\item RQ3: What is the impact of the number of quantiles on CQE's performance?

\end{itemize}

By exploring these questions, we seek to provide a holistic view of CQE's capabilities, its practical impact, and its potential for generalization across different recommendation contexts.

\subsection{Online Experiments (RQ1)}

To validate the real-world impact of our Conditional Quantile Estimation (CQE) framework, we conducted extensive online A/B tests on a major platform with hundreds of millions of daily active users, ensuring statistical significance. 

\subsubsection{Experiment Setup}

Users were randomly assigned to either the control or experimental group, with each group receiving at least 10\% of the daily user traffic. Each online A/B test ran for over a week, providing ample time for data collection and reliable result analysis.

Our CQE model utilizes an MLP network and is integrated into the ranking stage for watch time prediction. This CQE-based prediction is then combined with other existing prediction signals (such as like rate, video completion rate, etc.) to comprehensively rank the candidate videos. 

We evaluated the performance of the recommendation system using four key metrics:

\begin{enumerate}[leftmargin=*]
    \item \textbf{Average Watch Time per User}: This core metric directly measures user engagement by quantifying the average time users spend watching recommended videos.
    \item \textbf{Total Play Count}: This metric accounts for the cumulative number of video plays across all users, reflecting the frequency of user interactions with the recommended content.
    \item \textbf{Active Days per User}: This metric measures the number of days users engage with the platform, indicating user retention.
    \item \textbf{Active Users per Day}: This metric represents the number of unique users who interact with the platform, reflecting the system's ability to maintain and grow its user base.
\end{enumerate}

\subsubsection{\textbf{Experiment Results (RQ1)}}

\begin{table}[t] % 'h' for "here"
\centering
\caption{Conservative Estimation's enhancements in active days and active users. A boldface means a statistically significant result  (p-value $<$ 5\%).}
\label{tab:user_retention}
\begin{tabular}{cc}
\hline 
Active days &  Active Users  \\
\hline 
\textbf{+0.033\%}
& \textbf{+0.031\%} \\ 
\hline 
\end{tabular}
\end{table}

\begin{table}[t] % 'h' for "here"
\centering
\caption{Comparison between CQE Strategies and baseline online. A boldface means a statistically significant result (p-value $<$ 5\%).}
\label{tab:online_results}
\label{tab:online-experiments}
\begin{tabular}{ccc}
\hline
Strategy &  Watch Time &  Play Count \\
\hline
Conservative Estimation  & +0.008\% & \textbf{+0.346\%} \\
Dynamic Quantile Combination & \textbf{+0.106}\% & \textbf{+0.177}\% \\
Conditional Expectation & \textbf{+0.165}\% & -0.088\% \\ 
\hline
\end{tabular}
\vspace{-0.3cm}
\end{table}

Table \ref{tab:online_results} and Table \ref{tab:user_retention} summarize the performance of the CQE strategies relative to the baseline:

\textbf{Conservative Estimation (CSE):} CSE yielded a balanced improvement across all metrics. It achieved a modest 0.008\% increase in "Average Watch Time per User" while simultaneously boosting "Total Play Count" by 0.346\%. Furthermore, CSE led to a 0.033\% increase in active days and a 0.031\% growth in active users. Given the massive daily user base of hundreds of millions, even a tiny increase in active days/active users is considered statistically significant at the 0.02\% level. These results indicate that CSE successfully encourages users to interact with more videos, visit the platform more frequently, and maintain longer-term engagement.

\begin{figure}[t]
  \centering
   \includegraphics[width=0.9\linewidth]{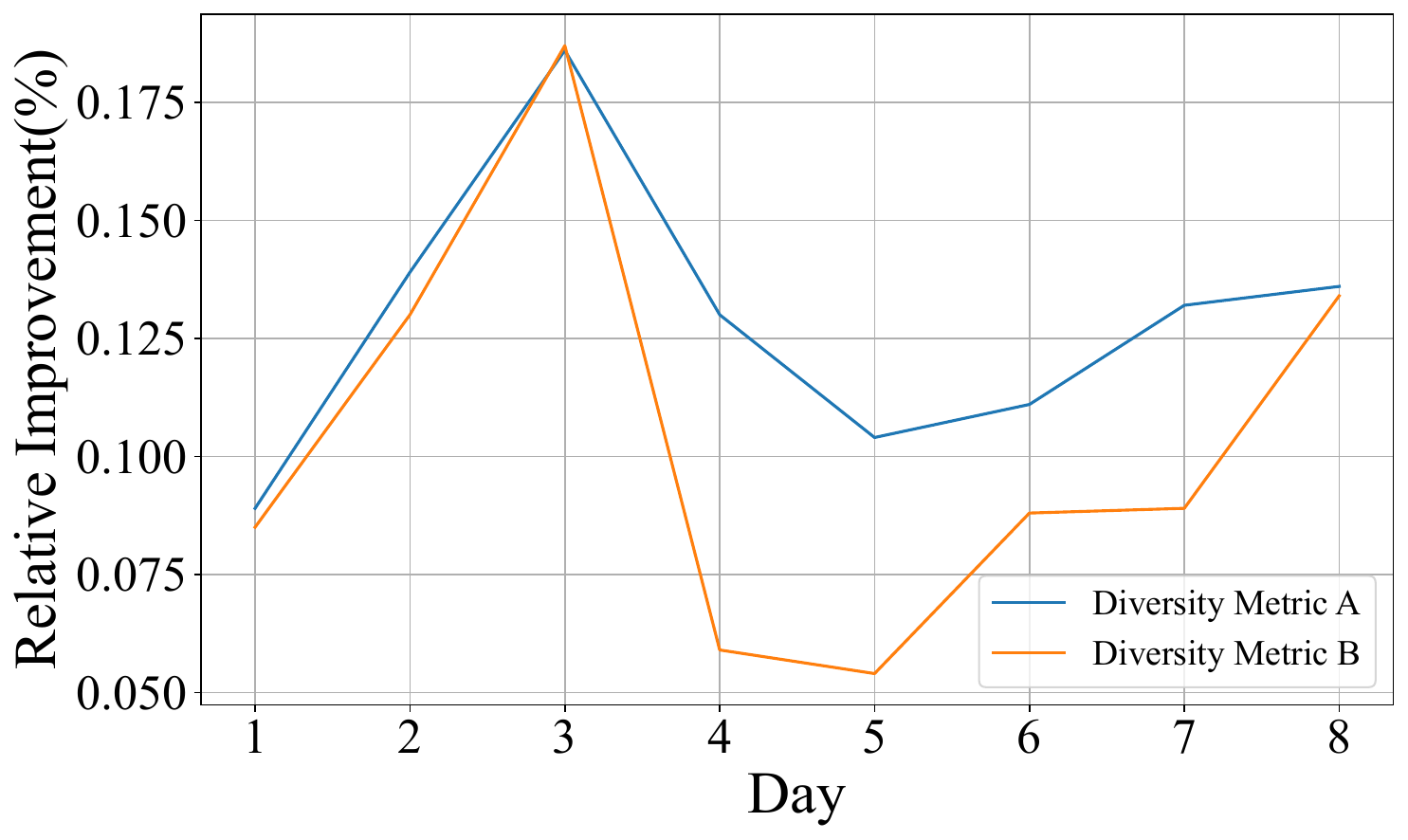}
   \vspace{-0.4cm}
  \caption{The diversity performances of CQE\textsubscript{DQC} compared with baseline online.}
  \label{fig:diversity}
\vspace{-0.3cm}
\end{figure}

\textbf{Dynamic Quantile Combination (DQC):} By setting the blending parameter k based on content novelty, DQC achieved improvements across both engagement and interaction metrics. It increased "Average Watch Time per User" by 0.106\% and "Total Play Count" by 0.177\%. Figure \ref{fig:diversity} illustrates that it also increased two core diversity metrics over the course of the experiment.

\textbf{Conditional Expectation (CDE):} The CDE methodology demonstrated a statistically significant increase of 0.165\% in the "Average Watch Time per User" metric. However, it showed a slight decrease of 0.088\% in "Total Play Count". This suggests that CDE effectively increased the depth of user engagement with individual videos, albeit at the cost of slightly reduced breadth of interaction.

These results collectively demonstrate the effectiveness of our CQE framework in improving various aspects of the recommendation system. Each strategy offers unique benefits:
\begin{itemize}[leftmargin=*]
\item CSE provides a balanced approach, improving all metrics and particularly excelling at encouraging broader platform interaction and user retention.
\item DQC offers a middle ground, improving both depth and breadth of engagement while  enhancing content diversity.
\item CDE excels at deepening user engagement with individual content pieces.
\end{itemize}

 The choice between these strategies would depend on specific platform goals, such as prioritizing deep engagement, broad interaction, user retention, or content diversity. Moreover, these strategies could potentially be combined or dynamically applied based on user segments or content types to optimize overall system performance.

\subsection{Distribution Characteristics Analysis}

To provide deeper insights into the nature of watch time uncertainty and validate our approach's applicability across diverse scenarios, we conducted additional analysis on the distribution characteristics of predicted watch times using online production data.

We defined the interquartile range (IQR) as $\frac{1}{N/2} \sum_{i=1}^{N/2} t_{\tau_{N-i+1}} - t_{\tau_i}$ to measure the spread of the conditional watch time distribution. Table~\ref{tab:iqr_performance} shows the relationship between IQR ranges and prediction accuracy metrics.

\begin{table}[t]
\caption{Prediction performance across different IQR ranges in online environment}
\label{tab:iqr_performance}
\centering
\begin{tabular}{cccc}
\hline
IQR Range & UAUC & MAE & Sample Ratio \\
\hline
$[0.0,0.1]$ & 0.721 & 0.105 & 0.077 \\
$[0.1,0.2]$ & 0.693 & 0.148 & 0.226 \\
$[0.2,0.4]$ & 0.631 & 0.197 & 0.543 \\
$[0.4,0.5]$ & 0.573 & 0.266 & 0.134 \\
$[0.5,1.0]$ & 0.536 & 0.305 & 0.020 \\
\hline
\end{tabular}
\end{table}

As shown in Table \ref{tab:iqr_performance}, the results reveal a clear pattern: samples with narrower distribution spreads (smaller IQR) consistently achieve better prediction performance. Notably, the majority of samples (54.3\%) fall within the medium IQR range $[0.2,0.4]$, while only 2\% of samples exhibit very wide distributions with IQR in $[0.5,1.0]$. This validates our approach's focus on modeling the full conditional distribution rather than a single point estimate, as the distribution spread varies significantly across different user-video pairs.

Furthermore, we analyzed how user activity levels affect prediction uncertainty in our online system, as shown in Table~\ref{tab:activity_performance}.

\begin{table}[t]
\caption{Prediction performance across user activity levels in online environment}
\label{tab:activity_performance}
\centering
\begin{tabular}{cccc}
\hline
User Activity Level & IQR & UAUC & MAE \\
\hline
Low Activity & 0.289 & 0.620 & 0.200 \\
Medium Activity & 0.274 & 0.634 & 0.195 \\
High Activity & 0.273 & 0.643 & 0.193 \\
\hline
\end{tabular}
\end{table}

These findings align with our theoretical understanding: (1) less active users exhibit higher prediction uncertainty (larger IQR); (2) prediction accuracy improves as user activity increases, likely due to richer historical data; and (3) even highly active users show inherent prediction uncertainty, underscoring the universal need for our distribution-aware approach regardless of user engagement level.

This empirical evidence from online data strongly supports our quantile-based framework and provides additional justification for the three inference strategies introduced in Section~\ref{sec: method}, each designed to handle different aspects of prediction uncertainty.

\subsection{Offline Experiments (RQ2 and RQ3)} \label{sec:offline}

 \begin{table*}[]
 \caption{Comparison of CQE with other methods across different backbones in the user interest prediction task. Additional results for the DCNV2 backbone are provided in Appendix \ref{sec:supA}. Boldface highlights the best-performing method for each backbone.}
  \label{tab:user-interest}
\begin{tabular}{cccccccccc}
\toprule
\multirow{2}{*}{Backbone} & \multirow{2}{*}{Method} & \multicolumn{4}{c}{KuaiRand}    & \multicolumn{4}{c}{WeChat}       \\
                          &                         & GAUC  & nDCG@1 & nDCG@3 & nDCG@5 & GAUC  & nDCG@1 & nDCG@3 & nDCG@5 \\ \midrule
\multirow{9}{*}{DeepFM}   & 
PCR + MSE & 0.592& 0.411& 0.465& 0.502& 0.537& 0.565& 0.572& 0.578  \\
& PCR + CE                   & 0.635 & 0.435  & 0.486  & 0.521  & 0.553 & 0.584  & 0.586  & 0.591  \\
                          & PCR + CQE\textsubscript{CDE}(Ours)     & 0.638 & 0.438 & 0.488 & 0.523 & \textbf{0.593} & \textbf{0.654} & \textbf{0.649} & \textbf{0.649}   \\  \cline{2-10} 
                          & WTG + MSE & 0.601& 0.417& 0.470& 0.506&0.537& 0.578& 0.579& 0.585
\\
                          & WTG     + CE           & 0.647 & 0.440  & 0.491  & 0.526  & 0.545 & 0.565  & 0.569  & 0.576  \\
                          & WTG + CQE\textsubscript{CDE}(Ours)     & 0.654 & 0.444 & 0.495 & 0.529 & 0.577 & 0.621 & 0.622 & 0.625   \\  \cline{2-10} 
                          & $\text{D}^2$Co + MSE & 0.606& 0.416& 0.473& 0.509& 0.528& 0.578& 0.577& 0.580
                          \\
                          & $\text{D}^2$Co + CE                   & 0.656 & 0.447  & 0.496  & 0.530  & 0.531 & 0.570  & 0.571  & 0.576  \\
                          & $\text{D}^2$Co + CQE\textsubscript{CDE}(Ours)    & \textbf{0.662} & \textbf{0.452} & \textbf{0.500} & \textbf{0.533} & 0.557 & 0.635 & 0.626 & 0.625   \\ \midrule
\multirow{9}{*}{AutoInt} 
& PCR + MSE &  0.595& 0.414& 0.467& 0.503& 0.536& 0.565& 0.572& 0.578 \\
& PCR + CE                    &      0.639 &      0.437    &   0.488      & 0.522      &      0.554 &     0.579   &    0.583    &     0.588   \\
                          & PCR + CQE\textsubscript{CDE}(Ours)     & 0.640 & 0.441 & 0.489 & 0.524 & \textbf{0.593} & \textbf{0.655} & \textbf{0.649} & \textbf{0.649}      \\ \cline{2-10} 
                          & WTG + MSE &  0.602& 0.419& 0.470 & 0.507& 0.536& 0.568& 0.573& 0.578 \\
                          & WTG     + CE                &     0.648  &    0.441    &   0.492     &   0.526     &  0.546     &  0.563      &   0.570     &   0.576     \\
                          & WTG + CQE\textsubscript{CDE}(Ours)     & 0.652 & 0.448 & 0.495 & 0.528 & 0.576 & 0.631 & 0.625 & 0.627     \\ \cline{2-10} & $\text{D}^2$Co + MSE & 0.610& 0.420& 0.474& 0.510& 0.527& 0.582& 0.580& 0.583
                          \\
                          & $\text{D}^2$Co    + CE                &     0.658  &   0.447     & 0.497       &   0.531     &    0.531   &   0.570     &    0.571    &     0.576   \\
                          & $\text{D}^2$Co + CQE\textsubscript{CDE}(Ours)    &    \textbf{0.661} & \textbf{0.455} & \textbf{0.500} & \textbf{0.534} & 0.557 & 0.635 & 0.626 & 0.624       \\ \bottomrule
\end{tabular}
\end{table*}

While our online A/B tests demonstrated the practical impact of CQE in real-world scenarios, offline experiments allow us to conduct more controlled and detailed analyses of our approach. Our offline experiments focus on two closely related tasks: watch time prediction and user interest prediction. Together, these tasks provide a comprehensive perspective for evaluating the effectiveness of the CQE framework in recommendation systems.

Watch time prediction directly captures the duration of user engagement with content, a key indicator of user involvement. However, predicting watch time alone may not fully capture user interest. Therefore, we introduce the user interest prediction task, accounting for video duration bias to offer a more refined measure of user interest. These two tasks complement each other: watch time prediction offers direct behavioral forecasting, while user interest prediction helps us understand the motivations behind these behaviors.

Following their publicly available code, we adopted the same network architecture and training configuration as TPM~\cite{TPM} for watch time prediction and $\text{D}^2$Co~\cite{D2Co} for user interest prediction.

\subsubsection{Watch-Time Prediction.} In this task, our primary objective is to accurately predict the duration of user watch time. 

\textbf{Datasets.} Following TPM~\cite{TPM}, we used two public datasets: Kuaishou (collected from Kuaishou App\footnote{https://kuairec.com/}) and CIKM16 (from CIKM16 Cup \footnote{https://competitions.codalab.org/competitions/11161}) for our experiments. 
While CIKM16 is primarily an e-commerce search query dataset, we include it to demonstrate the potential generalizability of our CQE approach across different recommendation contexts. The prediction of page dwell time in e-commerce shares similarities with video watch time prediction in terms of modeling user engagement duration, although we acknowledge the differences in content type and user behavior patterns.
In CIKM16 dataset, each item in the session is used as a single feature for input. Kuaishou dataset contains 7,176 users, 10,728 items, and 12,530,806 impressions; and CIKM16 dataset contains 310,302 sessions, and 122,991 items and the average length of each session is 3.981.

\textbf{Metrics.} We used two metrics to evaluate the model's performance: Mean Average Error (MAE) and XAUC~\cite{D2Q}.
\begin{itemize}[leftmargin=*]
    \item MAE: This metric is a typical measurement to assess the accuracy of the regression. Denote the prediction as $\hat y$  and the true watch time as $y$, 
 \begin{equation}
     \text{MAE} = \frac{1} {N} \sum_{i=1}^N\|\hat y_i - y\|.
 \end{equation}
\item XAUC: It evaluates if the predictions of two samples are in the same order as their true watch time.
It aligns well with the ranking nature of recommendation systems. In practice, the relative order of predictions often matters more than their absolute values, making XAUC particularly relevant to our research goals. 

\end{itemize}

\textbf{Baselines.} Six state-of-the-art methods for watch time prediction were selected for comparison, including WLR~\cite{youtube}, D2Q~\cite{D2Q}, OR~\cite{OR}, TPM~\cite{TPM}, DML~\cite{DML} and CREAD~\cite{CRAED}. The results of the first four methods are sourced from the TPM paper, while the latter two are based on our own reimplementations, as their official code is not publicly available.

\subsubsection{User Interest Prediction.} 
This task accounts for video duration bias to extract user interest from watch time. By doing so, we consider not just how long a user spent watching a video, but also how this time relates to the total video length, thus more accurately reflecting the user's true level of interest.

Following $\text{D}^2$Co \cite{D2Co}, in detail, we define the user interest for a given user-video pair $(u,v)$ as:
\begin{equation}
    x = \begin{cases} 
1, & \text{if } (d \leq 18s \text{ and } w = d) \text{ or } (d > 18s \text{ and } w > 18s); \\
0, & \text{otherwise};
\end{cases}
\end{equation}
where $d$ is video duration and $w$ is watch-time. We adopted the same training configuration as $\text{D}^2$Co and used the classical deep recommendation model DeepFM \cite{DFM}, the state-of-the-art recommendation model AutoInt \cite{AFI} and DCNV2\cite{DCNV2} as our backbone recommendation model.

\textbf{Datasets.} Following $\text{D}^2$Co, we leveraged two publicly available real-world datasets: WeChat\footnote{https://algo.weixin.qq.com/} and KuaiRand\footnote{http://kuairand.com/}. These datasets are sourced from prominent micro-video platforms, namely WeChat Channels and Kuaishou. WeChat dataset contains 20,000 users, 96,418 items, 7,310,108 interactions. This dataset, provided through the WeChat Big Data Challenge 2021, encompasses logs from WeChat Channels spanning a two-week period. KuaiRand dataset is a newly released sequential recommendation dataset collected from KuaiShou. As suggested in \cite{KuaiRand}, we utilized one of the subsets KuaiRand-pure in this study. It contains 26,988 users, 6,598 items, and 1,266,560 interactions.

\textbf{Metrics.} GAUC(Group Area Under Curve)~\cite{GAUC} and nDCG@k (normalized Discounted Cumulative Gain at rank k)~\cite{NDCG} are utilized as the evaluation metric of recommendation performance. 
\begin{itemize}[leftmargin=*]
    \item GAUC: this metric is calculated by weighted averaging the Area Under the ROC Curve (AUC) across different user groups, reflecting the model's ability to rank items accurately.
    \item nDCG@k: this metric measures the gain of a recommendation list based on the relevance of items and their positions up to the kth rank, offering insight into the quality of the top recommended items and their ordering.
\end{itemize}

\textbf{Baselines.} 
We used the weighted binary cross-entropy loss defined in $\text{D}^2$Co and Mean Squared Loss (MSE) as our baseline.  The binary cross-entropy loss is defined as 
\begin{equation}
 \mathcal L_{CE} =  -r \log [ \sigma ( f(x) )] -(1-r) \log [1 - \sigma ( f(x) ),
\end{equation}
where $\sigma$ is the sigmoid function and $r$ is the user’s interest defined by PCR(Play Completion Rate), WTG(Watch Time Gain) \cite{WTG} or $\text{D}^2$Co(Debiased and
Denoised watch time Correction)~\cite{D2Co}. Following $\text{D}^2$Co, in PCR and WTG, we treat all samples with less than 5 seconds of watch time as 0 values after calculating the value of labels. This can help remove the noise in watch time.

By default, we set the number of quantiles $N$ to 100. The value of $\tau_{low}$ is selected empirically from 0.2, 0.25, and 0.3. Similarly, the value of $\tau_{high}$ is selected empirically from 0.5, 0.6, 0.7, and 0.8.
\begin{table}
 \caption{Comparison between CQE and other approaches in watch-time prediction task. Boldface means the best-performed methods.}
 \label{tab:watch-time}
\begin{tabular}{ccccc}
\toprule
\multirow{2}{*}{Methods} & \multicolumn{2}{c}{Kuaishou} & \multicolumn{2}{c}{CIKM16} \\
                         & MAE           & XAUC         & MAE          & XAUC        \\ \midrule
WLR                      & 6.047         & 0.525        & 0.998        & 0.672       \\
D2Q                      & 5.426         & 0.565        & 0.899        & 0.661       \\
OR                       & 5.321         & 0.558        & 0.918        & 0.664       \\
TPM                      & 4.741         & 0.599        & 0.884        & 0.676       \\
DML                      & 4.455 & 0.607 & 0.829 & 0.693 \\
CREAD & 4.458 & 0.608 &	0.831 & 0.691 \\
CQE\textsubscript{CDE}(Ours)                 & \textbf{4.437}         & \textbf{0.610}        & \textbf{0.823}       & \textbf{0.694}      \\ \bottomrule
\end{tabular}
\end{table}

\subsubsection{Experimental Results} We summarize the results as follows:

\vspace{0.1cm} \label{sec:exp_backbone}
\noindent\textbf{Comparison between CQE\textsubscript{CDE} and other methods (RQ2):}
We compared the performances of different approaches in the watch-time prediction task, and the results are listed in Table \ref{tab:watch-time}. 
% Both TPM and  CQE\textsubscript{CDE}  outperform other methods in both MAE and XAUC metrics, thereby highlighting the significance of incorporating uncertainty into the models. 
Our approach demonstrates superior performance on both metrics compared to other mtehods, thereby emphasizing the advantages of employing quantile modeling technique.
Additionally, the consistent behavior between MAE and XAUC metrics also verifies the feasibility of the watch-time estimation serving as ranking metrics.
As for the user interest prediction task, we compare different frameworks across backbone models (DeepFM, AutoInt and DCNV2) and various label designs (PCR, WTG, and $\text{D}^2$Co) and present the results in Table \ref{tab:user-interest} and Appendix \ref{sec:supA}. Our proposed CQE\textsubscript{CDE} consistently outmatches the alternatives in all cases, indicating the robustness and effectiveness of CQE\textsubscript{CDE}.
In terms of the optimization framework, CE generally performs better than MSE, indicating the correctness of including the ordinal categorical information as guidance.
And CQE\textsubscript{CDE} can improve over CE across all the designs of user interest metrics (PCR, WTG, and $\text{D}^2$Co), which means that the proposed framework is generalizable to different label settings.

\begin{figure}[t]
  \centering
   \includegraphics[width=0.9\linewidth]{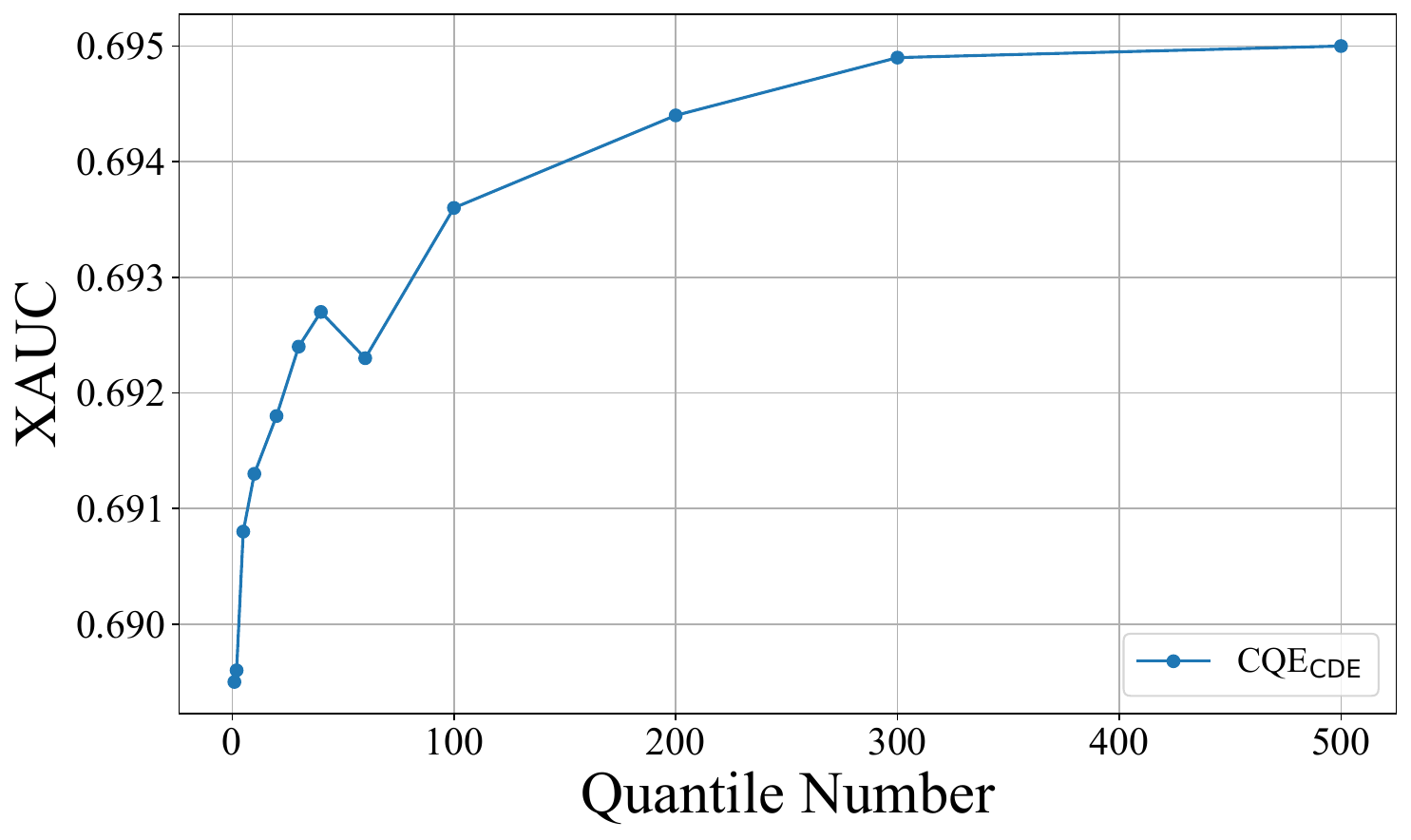}
   \vspace{-0.2cm}
  \caption{The performances of CQE\textsubscript{CDE} with various number of quantiles in CIKM16 Dataset for predicting watch time.}
  \Description{A woman and a girl in white dresses sit in an open car.}
  \label{fig:watch_time}
  \vspace{-0.2cm}
\end{figure}

\begin{figure}[t]
  \centering
  
  \includegraphics[width=0.9\linewidth]{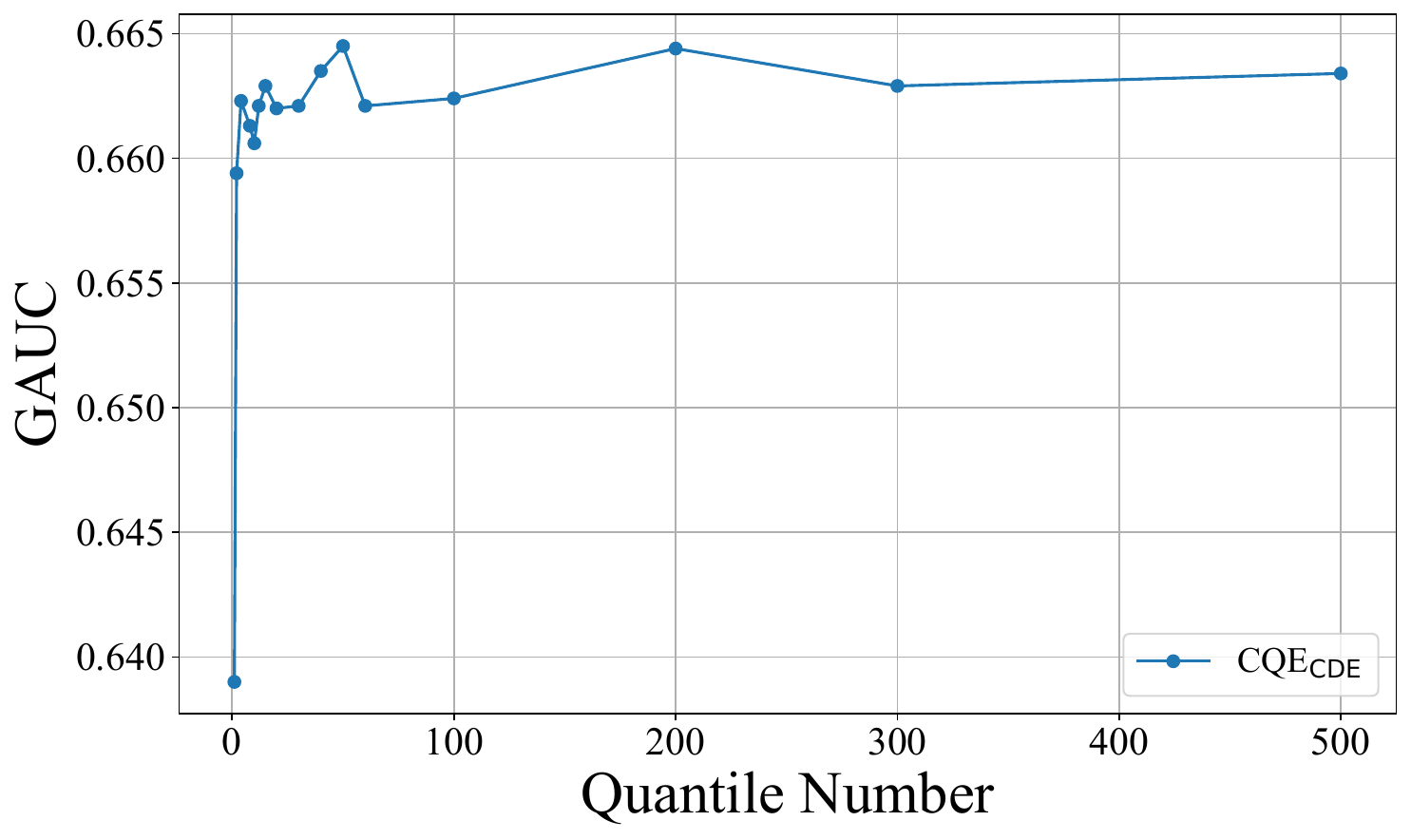}
  \vspace{-0.2cm}
  \caption{The performances of CQE\textsubscript{CDE} with various number of quantiles in KuaiRand Dataset for predicting user interest.}
    \vspace{-0.3cm}
  
  \label{fig:user_interest}
\end{figure}

\vspace{0.1cm}
\noindent\textbf{The Effect of Hyper-Parameters in CQE\textsubscript{CDE} (RQ3):}
To better investigate the characteristics of the proposed CQE framework, we further conduct an ablation study on the number of quantiles $N$ by varying its value from 1 to 500.
Theoretically, larger $N$ generates a more accurate approximation of the truth expectation and in turn achieves better recommendation performance in general.
This is attributed to the fact that more quantiles yield a distribution that closely mirrors the actual one.
As shown in Figure \ref{fig:watch_time}, model performance in the watch-time prediction task improves with an increasing number of predicted quantiles. For user interest prediction (Figure \ref{fig:user_interest}), performance remains relatively weak when the number of quantiles is below 10. Beyond this threshold, results fluctuate around 0.663, indicating that unlike watch-time prediction, a higher number of quantiles does not necessarily enhance performance. This discrepancy suggests a misalignment between the training objective and the user interest labels in the test set. Nevertheless, increasing $N$ generally improves prediction accuracy under the Conditional Expectation strategy.
% , but there may exists several bottlenecks that potentially limit the performance: 1) the design of the target label; 2) the data characteristics; and 3) the computational concerns.

In summary, our offline experiments comprehensively demonstrate the superiority of the CQE method in predicting user behavior and interest through these two complementary tasks. The watch time prediction task validates CQE's accuracy in direct behavioral prediction, while the user interest prediction task further proves CQE's ability to effectively capture more complex user preferences. The combination of these tasks not only validates the effectiveness of our method but also highlights the flexibility and adaptability of the CQE framework in addressing different yet related challenges in recommendation systems.

\section{Conclusion}
In this paper, we propose the Conditional Quantile Estimation (CQE) framework to model the uncertainty in watch-time prediction for short-video recommendation. Unlike traditional methods that estimate a single expected watch time, CQE captures the entire conditional distribution using quantile regression, allowing for more flexible and personalized recommendations. We further introduce three recommendation strategies—Conservative Estimation (CSE), Dynamic Quantile Combination (DQC), and Conditional Expectation (CDE)—tailored for different scenarios. Extensive offline experiments and large-scale online A/B testing demonstrate CQE’s effectiveness, leading to significant improvements in user engagement metrics such as active days and watch time.

For future work, we aim to explore dynamic quantile combination with the whole range of quantiles, integrate multi-modal features to improve content understanding, and investigate reinforcement learning-based optimization for automated strategy selection. Further research can also focus on  bias mitigation to enhance fairness in recommendation.

\bibliographystyle{ACM-Reference-Format}
\bibliography{sample-base}

\appendix

\section{Supplementary Results}\label{sec:supA}
Similar to our findings with DeepFM and AutoInt (Table \ref{tab:user-interest}), experiments with DCNV2 \cite{DCNV2} backbone (Table \ref{tab:other_backbone}) demonstrate CQE's consistent superior performance in user interest modeling.
\begin{table*}[]
 \caption{Comparison between CQE and other methods in user interest prediction task with backbone DCNV2.  Boldface means the best-performed methods.}
  \label{tab:other_backbone}
\begin{tabular}{cccccccccc}
\toprule
\multirow{2}{*}{Backbone} & \multirow{2}{*}{Method} & \multicolumn{4}{c}{KuaiRand}    & \multicolumn{4}{c}{WeChat}       \\
                          &                         & GAUC  & nDCG@1 & nDCG@3 & nDCG@5 & GAUC  & nDCG@1 & nDCG@3 & nDCG@5 \\ \midrule
\multirow{9}{*}{DCNV2}   & 
PCR + MSE & 0.601 &	0.428 &	0.481 &	0.510 &	0.542 &	0.569 &	0.571 &	0.577  \\
& PCR + CE   & 0.638 & 0.435 & 0.486 & 0.519 & 0.555 & 0.588 & 0.589 & 0.599                 \\
                          & PCR + CQE\textsubscript{CDE}(Ours)   & 0.643 & 0.437 & 0.492 & 0.523 & \textbf{0.595} & \textbf{0.654} & \textbf{0.650} & \textbf{0.651}
                          \\  \cline{2-10} 
                          & WTG + MSE & 0.601 & 0.418 & 0.469 & 0.508 & 0.539 & 0.571 & 0.577 & 0.587  \\
                          & WTG     + CE          & 0.649 & 0.443 & 0.495 & 0.528 & 0.558 & 0.560 & 0.569 & 0.579  \\
                          & WTG + CQE\textsubscript{CDE}(Ours)     & 0.652 & 0.449 & 0.497 & 0.530 & 0.586 & 0.637 & 0.638 & 0.640  \\ 
                          \cline{2-10} 
                          & $\text{D}^2$Co + MSE & 0.620 & 0.431 & 0.481 & 0.525 & 0.525 & 0.561 & 0.568 & 0.571
                          \\
                          & $\text{D}^2$Co + CE                 & 0.659 & 0.448 & 0.498 & 0.531 & 0.534 & 0.571 & 0.579 & 0.592  \\
                          & $\text{D}^2$Co + CQE\textsubscript{CDE}(Ours)    & \textbf{0.663} & \textbf{0.455} & \textbf{0.502} & \textbf{0.533} & 0.565 & 0.638 & 0.621 & 0.625   \\ 
 \bottomrule
\end{tabular}
\end{table*}

\section{Supplementary Code}\label{sec:supB}
We provide the implementation of the Pinball Loss function of Equation \ref{eq:pinball} in both PyTorch and TensorFlow.

\subsection{PyTorch Implementation}
\begin{lstlisting}[language=Python, caption=Pinball Loss in PyTorch]
import torch

def pinball_loss(y_true, y_pred, tau=0.5):
    error = y_true - y_pred
    return torch.mean(torch.maximum(tau * error, (tau - 1) * error))
\end{lstlisting}

\subsection{TensorFlow Implementation}
\begin{lstlisting}[language=Python, caption=Pinball Loss in TensorFlow]
import tensorflow as tf

def pinball_loss(y_true, y_pred, tau=0.5):
    error = y_true - y_pred
    return tf.reduce_mean(tf.maximum(tau * error, (tau - 1) * error))
\end{lstlisting}

This loss function is essential for quantile regression and serves as a core component of our CQE model.

\end{document}